# View-invariant Deep Architecture for Human Action Recognition using late fusion

Chhavi Dhiman, *Member IEEE,* Dinesh Kumar Vishwakarma, *Senior Member, IEEE*

*Abstract*— Human action Recognition for unknown views is a challenging task. We propose a view-invariant deep human action recognition framework, which is a novel integration of two important action cues: motion and shape temporal dynamics (STD). The motion stream encapsulates the motion content of action as RGB Dynamic Images (RGB-DIs) which are processed by the fine-tuned InceptionV3 model. The STD stream learns long-term view-invariant shape dynamics of action using human pose model (HPM) based view-invariant features mined from structural similarity index matrix (SSIM) based key depth human pose frames. To predict the score of the test sample, three types of late fusion (maximum, average and product) techniques are applied on individual stream scores. To validate the performance of the proposed novel framework the experiments are performed using both cross subject and cross-view validation schemes on three publically available benchmarks- NUCLA multi-view dataset, UWA3D-II Activity dataset and NTU RGB-D Activity dataset. Our algorithm outperforms with existing state-of-the-arts significantly that is reported in terms of accuracy, receiver operating characteristic (ROC) curve and area under the curve (AUC).

*Index Terms*— human action recognition, spatial temporal dynamics, human pose model, late fusion

## I. Introduction

HUMAN Action Recognition (HAR) in videos has gained a lot of attention in the pattern recognition and computer vision community due to its wide range of applications in intelligent video surveillance, multimedia analysis, human-computer interaction, and healthcare. Despite the efforts made in this domain, the performance of human action recognition in videos is still a challenging task due to the two main reasons: (1) the intra-class inconsistencies of an action due to different motion speeds, intensity illumination and viewpoints (2) the inter-class similarity of action due to similar set of human poses. The performance of the recognition system depends basically on how efficiently the relevant information is utilized. Due to the emergence of affordable depth maps with the Microsoft Kinect device, 3D features [1] [2] [3] based research, is proliferating speedily. Depth maps simplify the process of irrelevant background details removal and illumination variations largely. Hence, depth videos have turned out as a preferred solution [4] [5]to handle intra-class inconsistency due to intensity illumination and represent the fine details of the human pose.

Recently, Convolutional Neural Networks (CNN) [6] gained exceptional success to classify both images and videos [7] [8] [9] with their outstanding modeling capacity and ability to provide discriminative representations from raw images. It is observed from the previous works [10] [11] that appearance, motion and temporal information act as important cues to understand human actions in an effective manner [12]. The multi-stream architectures [13]: two streams [14] and three streams [15] have boosted the response of CNN based recognition systems, by jointly exploiting RGB and depth based appearance and motion content of actions. Optical flow [16] and dense trajectories [17] are used majorly to represent the motion of the object in videos. However, these approaches are not tailored to incorporate viewpoint invariance in action recognition. Dense trajectories are sensitive to camera viewpoints and do not include explicit human pose details during the action. Depth human pose can be useful to understand the temporal structure and global motion of human gait for more accurate recognition. Therefore, we propose a view invariant two-stream deep human action recognition framework, which is a fusion of Shape Temporal Dynamic (STD) stream and motion stream. STD stream learns the depth sequence of human poses as view-invariant Human Pose Model [18] (HPM) features over a period using Long Short Term Memory (LSTM) architecture. The motion stream encapsulates motion details of the action as Dynamic Images (DIs). The key contributions of the works are three folds as follows:

- A two-stream view invariant deep framework is designed for human action recognition using late fusion of motion stream and Spatial Temporal Dynamic (STD) stream that capture motion involved in an action sequence as Dynamic Images (DI) and human shape dynamics as view invariant Human Pose model (HPM) [18].
- To make identification more effective discriminant human shape dynamics is learnt for only key frames rather than entire video sequence using sequence of Bi-LSTM and LSTM models. It enhances the leaning ability of the framework.
- A compact representation of action video defined as DIs is learnt over inception blocks of fine-tuned inceptionV3 model to project the DIs in high dimensional feature space.

C. Dhiman and D.K. Vishwakarma are with Biometric Research Laboratory, Department of Information Technology, Delhi Technological University (Formerly Delhi College of Engineering), Bawana Road, Delhi-110042, India (e-mail: dvishwakarma@gmail.com).

Three strategies of late fusion: max(), avg() and multiply() are performed on two streams.
- To evaluate the performance of the proposed framework, experiments are conducted using cross subject and cross view validation schemes on three challenging public datasets such as - NUCLA multi-view dataset [19], UWA3D-II Activity dataset [20] and NTU RGB-D Activity dataset [21]. The obtained recognition accuracy is compared with the similar state-of-the-arts and exhibit superior performance.

The paper is organized as follows. In section II the related recent works are discussed highlighting their contribution to handling view invariance with different types of action representations using RGB, depth and RGB-depth modalities. In section III, the proposed framework is discussed in detail, and experimental results are reported in section IV in terms of recognition accuracy, ROC curves, and AUC. The work is concluded in section V.

## II. RELATED WORK

Deep learning based human action recognition solutions is proliferating with an added advantage one over another. Multi-stream deep architectures [8] [9] have surpassed the performances single stream deep state-of-the-arts [1] [16] due to the fact that such architectures are enriched with fusion of different types of action cues- temporal, motion, and spatial. The motion between frames is majorly defined as optical flow [16]. Extraction of optical flow is quite slow and often governs the total processing time of video analysis. Recent works [22] avoids optical flow computation by encoding the motion information as motion vectors and dynamic images. Temporal pooling appeared as an efficient solution to preserve the temporal dynamics of the action over a video. Various temporal pooling approaches were presented, in which pooling function is applied on temporal templates [23], on time series representation of spatial or motion action features [24], and ranking functions on video frames [25]. It is observed that all these approaches captured the temporal dynamics of the action over short time intervals.

CNNs are the powerful feature descriptors, for the given input. For video analysis, in particular, it is very crucial to decide the way video information is presented to CNN. Videos are treated as a stream of still images [26]. To utilize the temporal details of an action sequence, 2D filters are replaced by 3D filters [1] in CNNs while providing a sub-video of fixed length, or a pack of a short sequence of video frames into an array of images. However, the approaches above effectively apprehend the local motion details within a small window, but cannot preserve longer-term motion patterns associated with certain actions. Motion binary history (MBH) [27], motion history image (MHI) and motion energy image (MEI) [28], based static image representation of RGB-D action sequence attempted to preserve long-term motion patterns but the process of generation of this representation involves loss of information. It is difficult to represent the complex long-term dynamics of action in a compact form.

The mainstream literature listed above [5] [8] [9] [17] targeted action recognition from a common viewpoint. Such frameworks fail to produce a good performance for different viewpoint test samples. The viewpoint dependence of the framework can be handled by incorporating the view wise geometric details of the actions. Geometry-based action representations [17] [29] has improved the performance of view-invariant action recognition. Li et al. [30] proposed hanklets to capture view-invariant dynamic properties of short tracklets. Zhang et al. [31] defined continuous virtual paths by linear transformations of the action descriptors to associate action from dissimilar views. Where each point on the virtual path represents the virtual view. Rahmani and Mian [32] suggested a non-linear knowledge transfer (NKTM) model that mapped dense trajectory action descriptors to canonical views. These approaches [30] [31] [32] lacked spatial, hence shape/appearance details which are an important cue of an action. The work [33] aligned view-specific features in the sparse feature spaces and transferred the view-shared features in the sparse feature space to maintain a view specific and a common dictionary set separately. The publically available datasets cater action samples for a limited number of views either three or five. Liu et al. [34] introduced human pose model (HPM) learned with synthetically generated 3D human poses for 180 different views $\theta \in (0, \pi)$ that makes HPM [18] model rich with a wide range of views for a pose. Later, Liu et al. [17] fused the non-linear knowledge transfer model based view independent dense trajectory action descriptor with view-invariant HPM features. Integration of spatial details improved the performance of recognition significantly.

## III. PROPOSED METHODOLOGY

The proposed RGB-D based human action recognition framework is demonstrated in Fig. 1. The architecture is designed by learning both motion, view-invariant deep shape of the object over a period. The motion content of the action is encrypted as dynamic images (DI), and the concept of transfer learning is used to understand the action in RGB videos with the help of InceptionV3. Geometric details of the shape of the object during actions are extracted as view-invariant Human Pose Model (HPM) [18] features which are learned in a sequential manner using one Bi-LSTM, and one LSTM layer followed by dense, dropout and softmax layers. Two streams are combined using a late fusion concept to predict the action.

### A. Depth-Human Pose Model (HPM) based action descriptor

Depth shape representation of human pose preserves the information about relative positions of the body parts. In the proposed work, fine details of depth human pose irrespective of the viewpoint are represented as view-invariant HPM features. HPM model [18] has similar architecture to the AlexNet [35] but it is trained using synthetically generated multi-viewpoint action data from 180 viewpoints by fitting human models to the CMU motion capture data [36]. It makes HPM insensitive to the view variations in the actions. To preserve the temporal structure of action, HPM features are learned over LSTM sequential model.

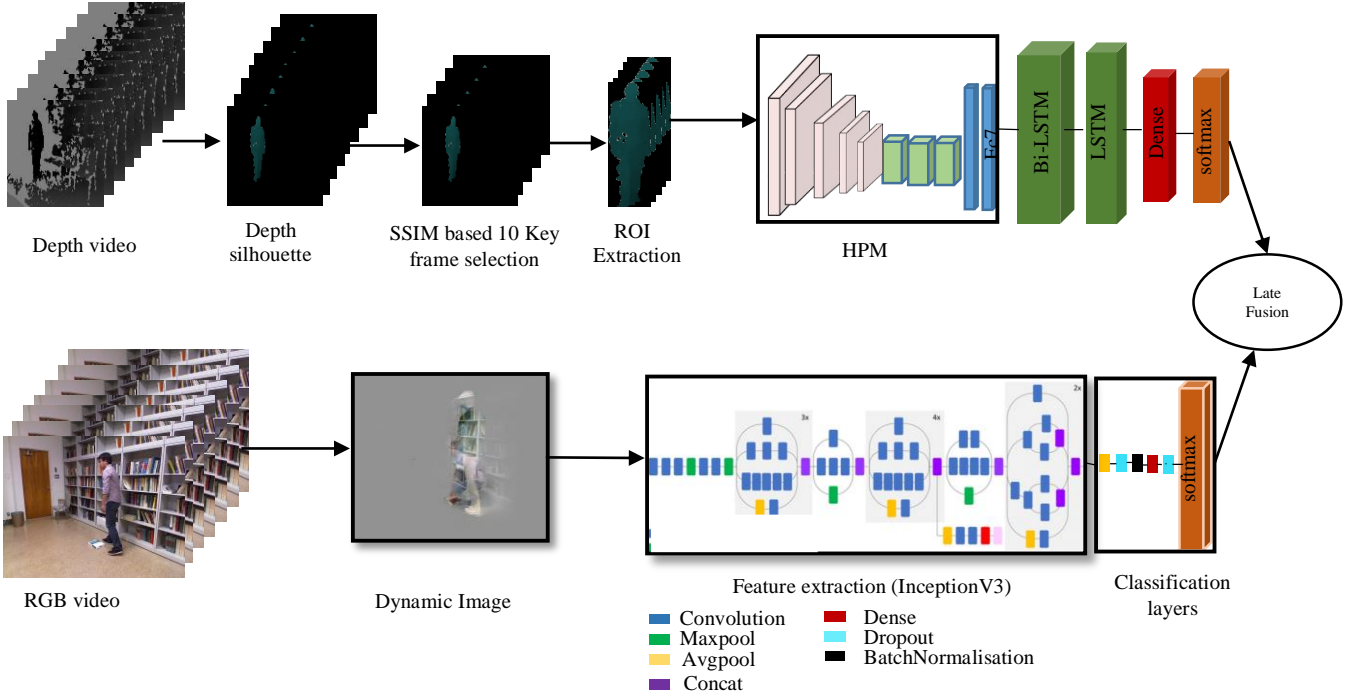

Fig. 1 Schematic Block Diagram of the proposed approach

1) *Self-Similarity Index Matrix (SSIM) based Key frame Extraction*

Initially a depth video with $n$ no. of depth frames $\{f_1, f_2 \ldots, f_n\}$, is pre-processed by morphological operations, to obtain depth human silhouette to reduce the background noise. The redundant information in the video is removed by selecting key pose frames based on the Structural Similarity Index Matrix (SSIM) [37]. It computes the global structural similarity index value and local SSIM map for two consecutive depth frames. If there are small changes in a human pose with time during an action structural similarity index ($\mathbb{SSI}$) value is high. For distinct human poses, the $\mathbb{SSI}$ value is small. Mathematically SSIM value is defined as below:

$$\mathbb{SSI}(f_i, f_{i+1}) = [\mathcal{L}(f_i, f_{i+1})^\alpha] \times [\mathbb{C}(f_i, f_{i+1})^\beta] \times [\mathcal{S}(f_i, f_{i+1})^\gamma] \quad (1)$$

where $\mathcal{L}(f_i, f_{i+1}) = \frac{(2*\vartheta_{f_i}*\vartheta_{f_{i+1}} + K_1)}{\vartheta_x^2 + \vartheta_y^2 + K_1}$, $\mathbb{C}(f_i, f_{i+1}) = \frac{(2*\sigma_x*\sigma_y + K_2)}{\sigma_x^2 + \sigma_y^2 + K_2}$, $\mathcal{S}(f_i, f_{i+1}) = \frac{(\sigma_{xy} + K_3)}{\sigma_x\sigma_y + K_3}$ where $\vartheta_x, \vartheta_y, \sigma_x, \sigma_y, \sigma_{xy}$ are the local means, variances and cross-variances for any two consecutive frames $f_i, f_{i+1}$ and $\mathcal{L}(.), \mathbb{C}(.)$ and $\mathcal{S}(.)$ are luminance, contrast and structural components of the pixels. Since depth images are not sensitive to luminance and contrast components, the exponents of $\mathcal{L}(.)$ and $\mathbb{C}(.)$ i.e. $\alpha, \beta$ are set to 0.5 and exponent of structural component $\mathcal{S}(.)$, $\gamma$ is set to 1. $\mathbb{SSI}$ value is computed for every two consecutive frames in a video and arranged in an ascending order with their respective frame numbers, in a vector $\Lambda$. First ten $\mathbb{SSI}$ values and corresponding frames numbers $i, i \in (1, n)$ are selected from the arranged vector $\Lambda$ as key frames. The salient information of each selected key frames is extracted as region of interest (ROI) and resized to $[227 \times 227]$ images to transform into view-invariant HPM features composed as $fc7$ layer $[10 \times 4096]$ feature vector using HPM [18] model.

2) *Model architecture and learning*

In this paper shape temporal dynamics (STD) stream is designed to describe the long term shape dynamics of the action with deep convolutional neural network (CNN) structure whose architecture is similar to [18] except that we have connected the last $fc7$ layer with a combination of Bidirectional LSTM and LSTM layers. The architecture of our CNN follows:
$Input(227,227) \rightarrow Conv(11,96,4) \rightarrow ReLU \rightarrow maxPool(3,2) \rightarrow Norm \rightarrow Conv(5,256,1) \rightarrow ReLU \rightarrow maxPool(3,2) \rightarrow Norm \rightarrow Conv(3,256,1) \rightarrow ReLU \rightarrow P(3,2) \rightarrow Fc6(4096) \rightarrow ReLU \rightarrow Dropout(0.5) \rightarrow Fc7(4096) \rightarrow BiLstm(512) \rightarrow Lstm(128) \rightarrow ReLU \rightarrow Fc(.) \rightarrow softmax(.)$

where $Conv(h, n, \mathbb{s})$ is a convolution layer with $h \times h$ kernel size, $n$ number of filters, $\mathbb{s}$ stride, $maxPool(h, \mathbb{s})$ is a max pooling layer of $h \times h$ kernel size and stride $\mathbb{s}$, $Norm$ is a normalization layer, $ReLU$ is a rectified linear unit, $Dropout(p)$ is Dropout layer with $(p)$ dropout ratio, $Fc(\mathbb{N})$ is a fully connected layer with $\mathbb{N}$ no. of neurons. $BiLstm(O)$ and $Lstm(O)$ are Bidirectional Long short term memory(LSTM) layer, and one directional LSTM later respectively with 'O' output shape. Bidirectional LSTM layer is trained with weight regularizer 0.001 and recurrent dropout of 0.55 with the true return sequence for Bidirectional LSTM layer. Softmax layer is attached in the end of the network. Last fully connected layer is designed with 10, 30, and 60 neurons as output shape for NUCLA, UWA3D, and NTU RGB-D dataset respectively according to number of classes in the datasets.

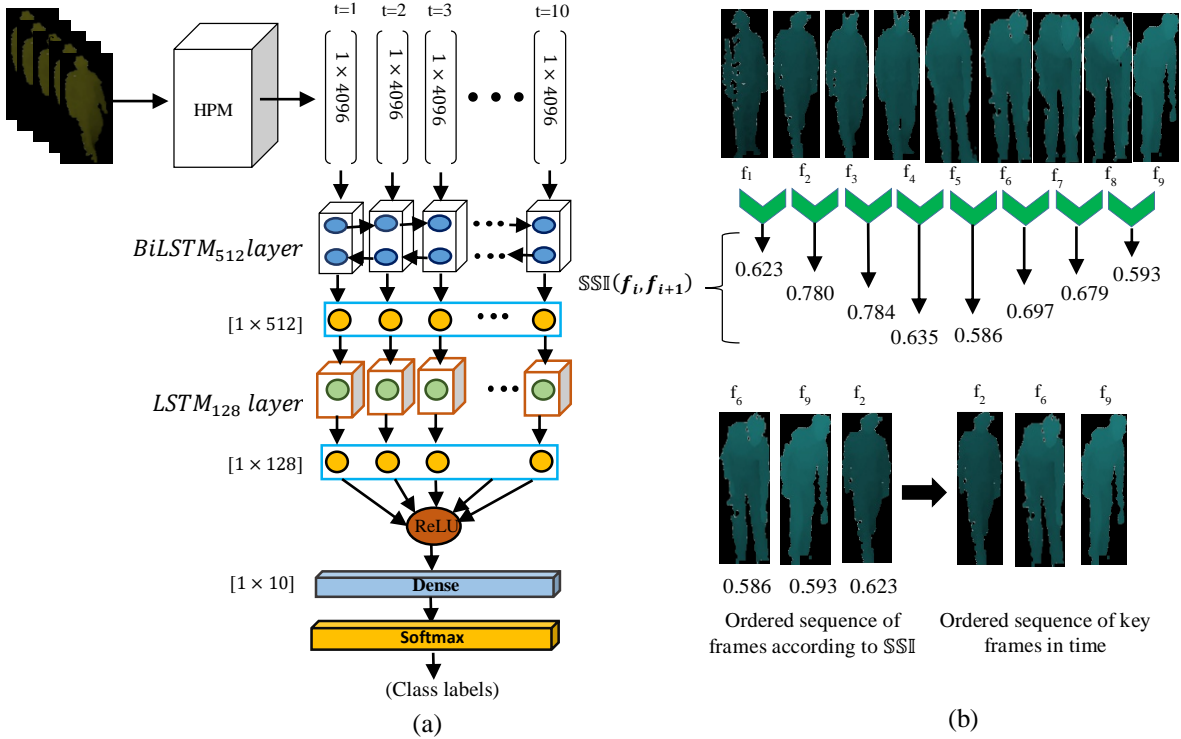

Fig. 2 (a) Shape Temporal Dynamics (STD) stream design (b) SSIM based key feature extraction procedure is demonstrated for only nine frames as a test case considering $\alpha = 0.5, \beta = 0.5, \gamma = 1$

The pre-trained HPM model is learned for view invariant synthetic action data for 399 types of human poses. Therefore, the proposed deep HPM based shape descriptor model is learned end to end with 80 epochs and 'Adam' optimizer. The $softmax$ layer will generate a probability vector $[1 \times n]$, where $n$ is no. of classes, that shows the belongingness of the test sample to all the classes of the dataset.

*B. RGB-Dynamic Image (DI) based action descriptor*

In this section appearance and dynamics of a video is represented in terms of dynamic images (DIs), which are later used to learn pre-trained inceptionV3 architecture according to the dynamics of the action sequence. DIs focus mainly on the salient objects and motion of the salient object by averaging away the background pixels and their motion patterns, by preserving long-term action dynamics. In comparison to other sequence invariant temporal pooling strategies [23] [25], ARP emphasize the order ($\tau$) of frame occurrence to extract complex long term dynamics of an action.

Construction of dynamic image depends on the ranking function that rank each frame in time axis. According to Fernando et al. [25], a video, i.e. $\{I_1, I_2, \ldots, I_N\}$ is represented as a ranking function $\varphi(I_t), t \in [1, N]$ where, $\varphi(.)$ function assigns a score $s$ to each frame $I_t$ at instance $t$ to reflect the rank of each frame. The time average of $\varphi(I_t)$ up to time $t$ is computed as $Q_t = \frac{1}{t}\sum_{i=1}^{t}\varphi(I_i)$ and $s(t|r) = <r, Q_t>$, where $r \in R^r$ is a vector of parameters. Score for each frame is computed in such a manner that $s(t_2|r) > s(t_1|r), t_2 > t_1$. For which vector $r$ is learned as a convex optimization problem using RankSVM [38]. The optimising equation is given as Eq. (2):

$$r^* = \partial(I_1, I_2, \ldots, I_N: \varphi) = argmin_r E(r), \quad (2)$$

where $E(r) = \frac{\lambda}{2}\|r\|^2 + \frac{2}{N(N-1)} \times \sum_{t_1 > t_2} \max\{0, 1 - s(t_2|r) + s(t_1|r)\}$

where $\partial(I_1, I_2, \ldots, I_N: \varphi)$ maps a sequence of $N$ number of frames to $r^*$, also termed as rank pooling function, that holds the information to rank all the frames in the video. The first term in objective function $E(r)$ is the quadratic regularised used in support vector machines. The second term is a hinge-loss that counts the number of pairs $t_2 > t_1$ are falsely ranked by the scoring function $s(.)$, if scores are not separated by at least unit margin i.e. $s(t_2|r) > s(t_1|r) + 1$.

In the proposed work, learning of the ranking function for dynamic images construction is accelerated by applying approximate rank pooling (ARP) [39]. It involves simple linear operations at pixel level, over the frames to rank them, which is extremely efficient and simple for fast computation. ARP approximates the rank pooling procedure by using gradient-based optimization in Eq. (1) as follow:

For $r = \vec{0}, r^* = \vec{0} - \eta \nabla E(r)|_{r=\vec{0}}$ for any $\eta > 0$,
where $\nabla E(r) \propto \sum_{t_2 > t_1} \nabla \max\{0, 1 - s(t_2|r) + s(t_1|r)\}|_{d=\vec{0}}$

$$\nabla E(r) = \sum_{t_2 > t_1} \nabla < r, Q_{t1} - Q_{t2} > = \sum_{t_2 > t_1} Q_{t1} - Q_{t2}$$

$$r^* \propto \sum_{t_2 > t_1} \left[\frac{1}{t_2}\sum_{i=1}^{t_2}\varphi_i - \frac{1}{t_1}\sum_{j=1}^{t_1}\varphi_j\right] = \sum_{t=1}^{T}\gamma_t \varphi_t \quad (3)$$

where $\gamma_t = 2(T - t + 1) - (T + 1)(h_t - h_{t-1})$, and $h_t = \sum_{i=1}^{t}\frac{1}{t}$ is the $t^{th}$ harmonic number, $h_0 = 0$. Hence, rank-pooling function is re-written as:

$$\hat{\partial}(I_1, I_2, \ldots, I_N: \varphi) = \sum_{t=1}^{T}\gamma_t \varphi(I_t) \quad (4)$$

$$\gamma_1 \rightarrow \quad f_1 \oplus \quad f_2 \oplus \quad f_3 \quad \cdots \quad f_{N-1} \oplus \quad f_N$$

$$\gamma_1 \rightarrow \frac{2*\mathbf{1}-N-1}{1} \quad \frac{2*\mathbf{2}-N-1}{2} \quad \frac{2*\mathbf{3}-N-1}{3} \quad \cdots \quad \frac{2*(\mathbf{N-1})-N-1}{(N-1)} \quad \frac{2*\mathbf{N}-N-1}{N}$$

$$\gamma_2 \rightarrow \quad \frac{2*\mathbf{2}-N-1}{2} \quad \frac{2*\mathbf{3}-N-1}{3} \quad \cdots \quad \frac{2*(\mathbf{N-1})-N-1}{(N-1)} \quad \frac{2*\mathbf{N}-N-1}{N}$$

$$\gamma_3 \rightarrow \quad \quad \frac{2*\mathbf{3}-N-1}{3} \quad \cdots \quad \frac{2*(\mathbf{N-1})-N-1}{(N-1)} \quad \frac{2*\mathbf{N}-N-1}{N}$$

$$\gamma_{N-1} \rightarrow \quad \frac{2*(\mathbf{N-1})-N-1}{(N-1)} \quad \frac{2*\mathbf{N}-N-1}{N}$$

$$\gamma_N \rightarrow \quad \frac{2*\mathbf{N}-N-1}{N}$$

Fig. 3: computation of $\gamma_t$ parameter for fixed video length $N$; numbers in red show the dependency of $\gamma_t$ on consecutive video frames $\in (i,N)$

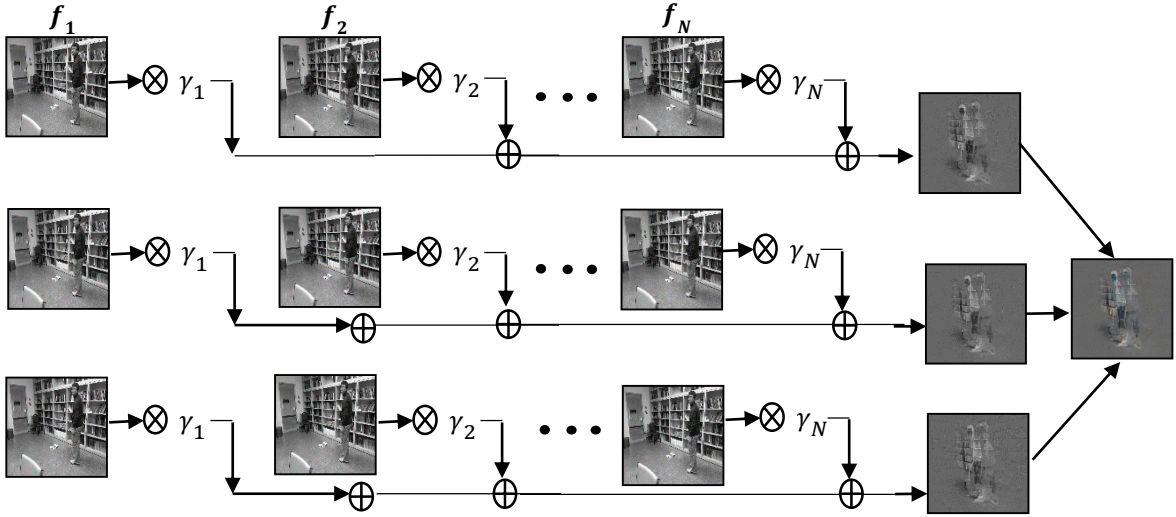

Fig. 4: Dynamic image formation using Approximate Rank Pooling (ARP) [48]. First row: R-channel, Second row: G-channel, Third row: B-channel of each RGB video frame.

Therefore, ARP can be defined as a weighted sum of sequential video frames. The weights $\gamma_t, t\epsilon[1,N]$ are pre computed for a fixed length video, using Eq. (3) as shown in Fig. 3. While computing $\gamma_n$, as defined in eq. (3), the order of occurrence of all the frames, for time $t \geq n$, are considered by computing a weight for each frame $\frac{2*i-N-1}{i}$, where $i \in [n,N]$. The computed weight value for each considered frame is summed to obtain single value of $\gamma_n$. Therefore, rank-pooling function can be directly defined by using individual frame features $\varphi(I_t)$ and $\gamma_t = 2(T-t+1)$ as a linear function of time $t$, instead of computing the intermediate average feature vectors $Q_t$ per frame to assign the score to rank the frames. The procedure of Approximate Rank Pooling (ARP) is shown in Fig. 4. Where each video frame is multiplied with the corresponding computed, $\gamma_t$ weight i.e. $f_1$ is multiplied with $\gamma_1$ for every channel separately. R, G, and B channels of the dynamic image is obtained as weighted sum of R, G, and B - channels of each video frame respectively. The size of the DI, so obtained, is same as original frame. To compute view invariant motion features of the action, the constructed DIs are passed through the motion stream as shown in Fig. 5, which is a combination of InceptionV3 architecture convolution layers followed by set of classification layers i.e. $Global\ Average\ Pooling\ Layer2D()$, $BatchNormalisation(), dropout(0.3)$, $dense(512,'Relu'), dropout(0.5), and\ Dense(10,'softmax')$ layers. Convolution features with vector shape of $8 \times 8 \times 2048$ is received as high dimensional representation of input image using pre-trained InceptionV3 model. Batchnormalisation layer is used to maintain the internal covariate shift of hidden units' values to be minimal after 'ReLU' activations in $dense(512,'Relu')$ layer. Combination of Batchnormalisation and Dropout layer helped to handle the overfitting phenomena together without minimal loss of dropouts rather than only depending on dropout layer resulting in larger loss of weights. The layers of the motion stream are trained end to end for multiview datasets to update the weights of the InceptionV3 convoultion layers according to training samples. The best trained model weights so obtained for the highest achieved validation accuracy are used for testing of the sample to achieve the high recognition rate irrespective of view variations.

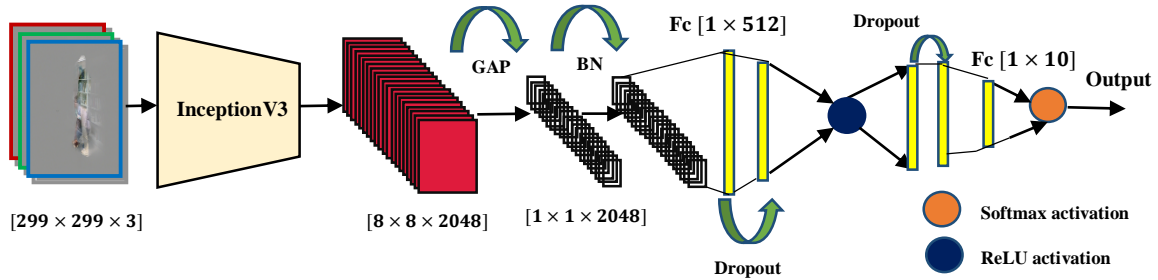

Fig. 5: Layer Structure of the Motion Stream, GAP: Global Average Pooling, BN: Batch normalization.

## IV. EXPERIMENTAL RESULTS

To validate the performance of the proposed framework for view-invariant human action recognition, three publically available NUCLA multi-view action 3D dataset, UWA3D Depth dataset and NTU RGB-D activity datasets are used.

### B. NUCLA multi-view action 3D Dataset

The Northern-UCLA multi-view RGB-D dataset [19] is captured by Jiang Wang and Xiaohan Nie in UCLA simultaneously from three different viewpoints using Kinect v1. The dataset covers 10 action categories performed by 10 subjects: (1) pick up with one hand, (2) pick up with two hands, (3) drop trash, (4) walk around, (5) sit down, (6) stand up, (7) donning, (8) doffing, (9) throw, (10). This dataset is very challenging because many actions share the same "walking" pattern before and after the actual action is performed. To handle this inter-class similarity SSIM based ten depth key frames are selected and processed as 3D-HPM shape features. Moreover, some actions such as "pick up with on hand" and "pick up with two hands" are difficult to discriminate from different viewpoints. For cross view validation two views are used for training, and third view is used for testing. For cross-subject validation, test samples are selected irrespective of viewpoint. The sample frames of the datasets are shown in Fig. 6(a).

### C. UWA3D Multi view Activity-II Dataset

UWA3D multi-view activity-II dataset [20] is a large dataset which covers 30 human actions performed by ten subjects and recorded from 4 different viewpoints at different times using the Kinect v1 sensor. The 30 actions are:(1) one hand waving, (2) one hand punching, (3) two hands waving, (4) two hands punching, (5) sitting down, (6) standing up, (7) vibrating, (8) falling down, (9) holding chest, (10) holding head, (11) holding back, (12) walking, (13) irregular walking, (14) lying down, (15) turning around, (16) drinking, (17) phone answering, (18) bending, (19) jumping jack, (20) running, (21) picking up, (22) putting down, (23) kicking, (24) jumping, (25) dancing, (26) moping floor, (27) sneezing, (28) sitting down (chair), (29) squatting, and (30) coughing. The four viewpoints are: (a) front, (b) left, (c) right, (d) top. The major challenge of the dataset lies in the fact that large number of action classes are not recorded simultaneously resulting in intra-action differences besides viewpoint variations. The dataset also contains self-occlusions and human-object interactions in some videos. Sample images of UWA3D dataset from four different viewpoints are shown in Fig. 6(b).

### D. NTU RGB-D Human Activity Dataset

NTU RGB+D action recognition dataset [21] is a large-scale RGB-D Dataset for human activity analysis captured by 3 Microsoft Kinect v.2 cameras placed at three different angles: $-45^0, 0^0, 45^0$, simultaneously. It consists of 56,880 action samples including RGB videos, depth map sequences, 3D skeletal data, and infrared videos for each sample. The dataset consists of 60 types (50 single person actions and 10 two-person interactions) of actions performed by 40 subjects repeated twice facing the left or right sensor respectively. The height of the sensors and their distances to the subject performing an action were further adjusted to get more viewpoint variations. This makes the NTU RGB-D dataset a largest and most complex cross-view action dataset of its kind to date. RGB and depth sample frames of NTU RGB-D dataset are shown in Fig. 6(c). The resolution of RGB videos and depth maps is 1920×1080 and 512×424 respectively. We follow the standard cross subject and cross view evaluation protocol in the experiments, as specified in [21]. Under cross subject evaluation protocol, out of 40 subjects 20 subjects are selected for training and 20 subjects for testing. Under cross subject evaluation protocol, out of 40, 20 subjects are selected for training and 20 subjects for testing. Under cross view evaluation protocol, view 2 and view 3 are used as training views and view 1 is used as test view.

In the experiments, both motion stream and STD stream of the proposed deep framework are pre-trained end-to-end independently. For testing phase, the best-trained model is selected based on highest validation accuracy achieved. Under cross view validation scheme one view is used as test view and rest all views are used for training. In the training phase, the training samples are split in training samples and validation samples using 80-20 splitting strategy and Adaptive Moment Estimation (Adam) optimizer is used with (epochs, batch size, learning rate of the Adam optimizer) as (80, 10, 0.0002). In the testing phase, the scores obtained from each stream for each test sample are fused using three late fusion mechanisms: maximum, average and product. The obtained performance of our approaches for cross subject and cross view validation scheme for NUCLA multi-view dataset and UWA3D II activity dataset is provided in Table I, II and III highlighting the obtained highest accuracy of the proposed framework for each dataset. Where results are described in terms of motion stream, STD stream and proposed hybrid approach which stands for [DI_InceptionV3], [HPM_LSTM] and [HPM_LSTM + DI_InceptionV3] respectively.

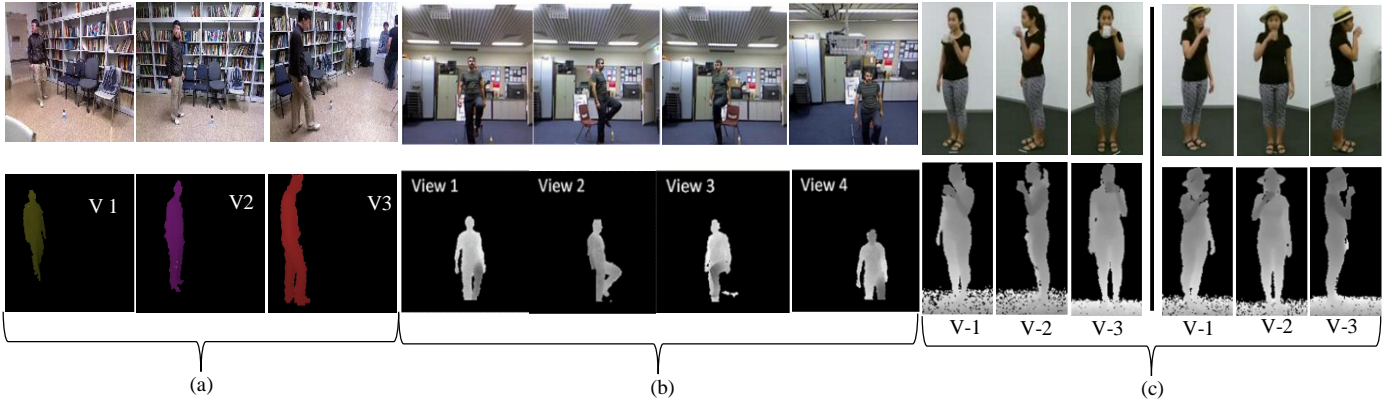

Fig. 6. Sample Images of (a) NUCLA multi-view 3D Action Dataset (b) UWA3D II Multi-View Action Dataset and (c) NTU RGB-D Human Activity Dataset. Left group of images in the Fig. 4(c) are recorded when the subject face camera sensor C-3, and right group of images are recorded when the subject face camera sensor C-2

It is observed that for both cross view and cross-subject validation, late fusion scheme has produced remarkable results. The small variation in the accuracy for different views exhibits the view invariance property of the proposed framework. Whereas, in other state-of-the-arts [18] [1] [34], accuracy of the presented frameworks vary from view to view in the range of 10% that shows these state-of-the-arts are sensitive to different views.

TABLE I
CROSS SUBJECT VALIDATION RESULTS FOR NUCLA AND UWA3D II MULTI-VIEW ACTION 3D DATASET AND NTU RGB-D ACTIVITY DATASET

| Dataset Method | | NUCLA dataset | UWA3D II dataset | NTU RGB-D dataset |
|---|---|---|---|---|
| Motion stream | | 93% | 82.6% | 62% |
| STD stream | | 76% | 73.5% | 68.3 |
| Proposed Hybrid Approach | Max | 83% | 81.8% | 71.6 |
| | Avg | 84.5 % | 79.6% | 75.7 |
| | Mul. | **87.3%** | **85.2%** | **79.4%** |

TABLE II
CROSS VIEW VALIDATION RESULTS FOR NUCLA MULTI-VIEW ACTION 3D DATASET

| Training/ Test View | | [1,2]/3 | [1,3] / 2 | [2,3]/ 1 | Mean |
|---|---|---|---|---|---|
| Motion stream | | 86.29 | 76.42 | 70.6 | 77.77 |
| STD stream | | 58.88 | 73.67 | 63.83 | 65.46 |
| Hybrid | Max | 83.08 | 79.7 | 78.72 | 80.5 |
| | Avg | **87.15** | 78.38 | 73.50 | 79.67 |
| | Mul. | 84.58 | **88.21** | **83.36** | **85.38** |

The comparison of the recognition accuracy is shown in Table IV, V, and VI for NUCLA, UWA3DII and NTU RGB-D dataset respectively. It is observed the novel integration of motion stream and STD stream of the proposed method has outperformed the recent works HPM_TM [18], HPM_TM+DT [17], NKTM [40]. Interestingly, our method achieves 91.3% and 83.6% average recognition accuracy which is about 9% and 10.86% higher than the nearest competitor HPM_TM+DT [17] when view 1 is considered for test view for both UWA3D II activity dataset and NUCLA dataset. However, the obtained classification accuracy for NTU RGB-D dataset is not as good as obtained from other datasets due to the large variety of number of samples and their complexity in NTU RGB + D dataset. Viewpoint and large intra-class variations make this dataset very challenging. The performance of other work [41], Table VI, is comparatively better than the proposed framework for NTU RGB-D Activity dataset. It utilized the skeleton joints based action features to make prediction.

However, the novel integration of motion stream and STD stream using late fusion has boosted recognition accuracy for all three multi-view datasets verified as ROC curves and AUC in Fig. 7 for individual test view of each multi-view dataset. From where it can be easily visualised that the hybrid approach based ROC curves are showing superior performance than the individual motion stream and STD stream based classification results which supports the fact that the fusion of the scores of two streams has resulted in increase in correct selection of true samples thereby improved true positive rate (TPR). At the same time, AUC values of the ROC curves help to understand and compare the ROC curves in a clearer way when they cross each other or nearly close to each other.

TABLE III
CROSS VIEW VALIDATION RESULTS FOR UWA3D MULTI VIEW ACTIVITY-II DATASET

| Training View | [v1,v2] | | [v1, v3] | | [v1,v4] | | [v2,v3] | | [v2,v4] | | [v3,v4] | | mean |
|---|---|---|---|---|---|---|---|---|---|---|---|---|---|
| Test View | v3 | v4 | v2 | v4 | v2 | v3 | v1 | v4 | v1 | v3 | v2 | v4 | |
| Motion Stream | 87.4 | 81.2 | 78.1 | 85.5 | 73.9 | 79.4 | 82.6 | 73.1 | 81.6 | 72.4 | 83.5 | 81.1 | 79.98 |
| STD stream | 62.1 | 73.5 | 69.6 | 79.6 | 65.4 | 75.9 | 64.3 | 69.5 | 66.3 | 69.8 | 78.6 | 68.8 | 70.2 |
| Proposed Hybrid Approach | 86.6 | **85.3** | 81.8 | 86.5 | 78.3 | 82.8 | 85.1 | 83.6 | 85.1 | 81.2 | 85.3 | 82.3 | 83.65 |
| | 73.2 | 78.8 | 75.4 | 81.3 | 79.9 | 81.4 | 79.4 | 77.3 | 79.4 | 80.9 | 84.1 | **84.2** | 79.6 |
| | **88.2** | 84.3 | **82.6** | **88.6** | **80.5** | **83.2** | **88.9** | **84.6** | **93.9** | **85.2** | **91.2** | 83.0 | **86.18** |

TABLE IV
COMPARISON OF ACTION RECOGNITION ACCURACY (%) ON NUCLA MULTI-VIEW ACTION 3D DATASET

| Train-Test View Methods | Data Type | [1,2]/ 3 | [1,3] /2 | [2,3] /1 | Mean |
|---|---|---|---|---|---|
| CVP [31] | RGB | 60.6 | 55.8 | 39.5 | 52 |
| nCTE [42] | RGB | 68.6 | 68.3 | 52.1 | 63 |
| NKTM [40] | RGB | 75.8 | 73.3 | 59.1 | 69.4 |
| HOPC+STK [20] | Depth | 80 | - | - | - |
| HPM_TM [18] | Depth | 92.2 | 78.5 | 68.5 | 79.7 |
| HPM_TM+DT [17] | RGBD | 92.9 | 82.8 | 72.5 | 82.7 |
| Motion Stream | RGB | 86.29 | 79.7 | 70.6 | 77.77 |
| STD stream | Depth | 58.8 | 73.67 | 63.83 | 65.46 |
| Proposed Hybrid Approach | RGBD | **84.58** | **88.21** | **83.36** | **85.38** |

TABLE V
COMPARISON OF ACTION RECOGNITION ACCURACY (%) UWA3D MULTI-VIEW ACTIVITY-II DATASET

| Train-Test View Methods | Data Type | [v1,v2] | | [v1, v3] | | [v1,v4] | | [v2,v3] | | [v2,v4] | | [v3,v4] | | Mean |
|---|---|---|---|---|---|---|---|---|---|---|---|---|---|---|
| | | v3 | v4 | v2 | v4 | v2 | v3 | v1 | v4 | v1 | v3 | v1 | v2 | |
| DT [43] | RGB | 57.1 | 59.9 | 60.6 | 54.1 | 61.2 | 60.8 | 71 | 59.5 | 68.4 | 51.1 | 69.5 | 51.5 | 60.4 |
| C3D [1] | RGB | 59.5 | 59.6 | 56.6 | 64 | 59.5 | 60.8 | 71.7 | 60 | 69.5 | 53.5 | 67.1 | 50.4 | 61 |
| nCTE [42] | RGB | 55.6 | 60.6 | 56.7 | 62.5 | 61.9 | 60.4 | 69.9 | 56.1 | 70.3 | 54.9 | 71.7 | 54.1 | 61.2 |
| NKTM [32] | RGB | 60.1 | 61.3 | 57.1 | 65.1 | 61.6 | 66.8 | 70.6 | 59.5 | 73.2 | 59.3 | 72.5 | 54.5 | 63.5 |
| R-NKTM [40] | RGB | 64.9 | 67.7 | 61.2 | 68.4 | 64.9 | 70.1 | 73.6 | 66.5 | 73.6 | 60.8 | 75.5 | 61.2 | 67.4 |
| HPM(RGB+D)_Traj [34] | RGBD | 85.8 | 89.9 | 79.3 | 85.4 | 74.4 | 78 | 83.3 | 73 | 91.1 | 82.1 | 90.3 | 80.5 | 82.8 |
| HPM_TM+DT [17] | RGBD | 86.9 | 89.8 | 81.9 | 89.5 | 76.7 | 83.6 | 83.6 | 79 | 89.6 | 82.1 | 89.2 | 83.8 | 84.6 |
| Motion Stream | RGB | 87.4 | 81.2 | 78.1 | 85.5 | 73.9 | 79.4 | 82.6 | 73.1 | 81.6 | 72.4 | 83.5 | 81.1 | 79.98 |
| STD stream | Depth | 62.1 | 73.5 | 69.6 | 79.6 | 65.4 | 75.9 | 64.3 | 69.5 | 66.3 | 69.8 | 78.6 | 68.8 | 70.2 |
| Proposed Hybrid Approach | RGBD | 86.6 | **85.3** | 81.8 | 86.5 | 78.3 | 82.8 | 85.1 | 83.6 | 85.1 | 81.2 | 85.3 | 82.3 | 83.65 |
| | | 73.2 | 78.8 | 75.4 | 81.3 | 79.9 | 81.4 | 79.4 | 77.3 | 79.4 | 80.9 | 84.1 | **84.2** | 79.6 |
| | | **88.2** | 84.3 | **82.6** | **88.6** | **80.5** | **83.2** | **88.9** | **84.6** | **93.9** | **85.2** | **91.2** | 83.0 | **86.18** |

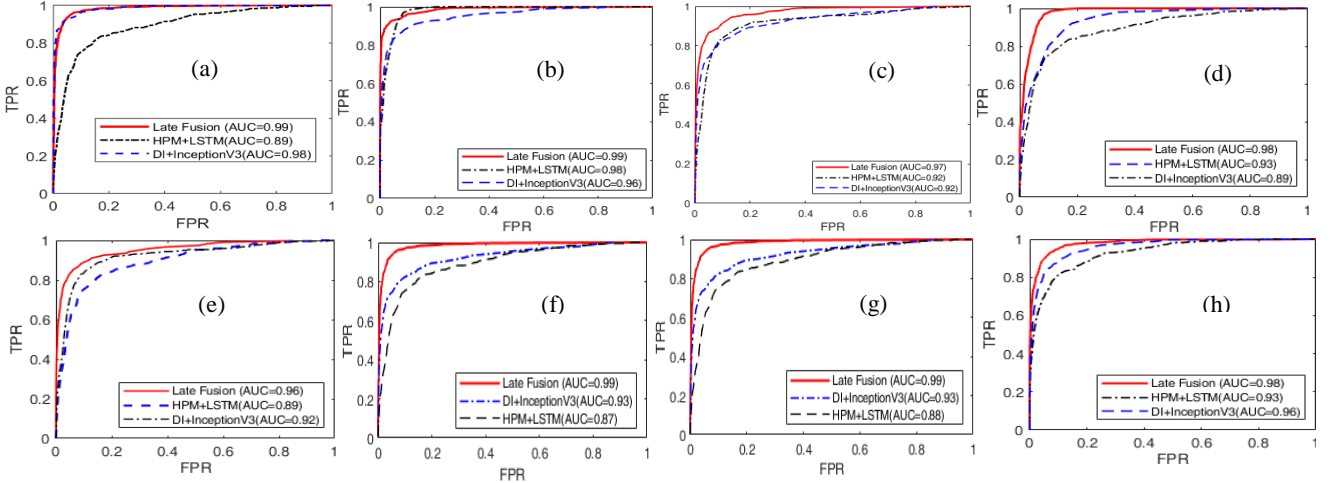

Fig. 7: Performance evaluation of the proposed framework for NUCLA multi-view dataset (a)-(c), UWA3D dataset (d)-(g) and NTU RGB-D Activity dataset (h) in terms of ROC curve and area under the curve (AUC).

*Computation time:* Our technique outperforms the current cross-view action recognition methods on multi-view NUCLA, UWA3D and NTU RGB-D Activity Dataset by fusing motion stream and view-invariant Shape Temporal Dynamics (STD) stream information. Therefore, the proposed two stream deep architecture not only perform proficiently but also time efficient compared to existing cross-view action recognition techniques. The experiments are performed on a single 8GB NVIDIA GeForce GTX 1080 GPU system. It does not demand computationally expensive training and testing phases, as shown in Table VII. The major reason behind lesser computation cost involved in training and testing phase is the compact and competent representation of action. In motion stream, the entire video sequence is represented by a single DI and STD stream process the key human pose depth frame instead of all the frames in the action sequence.

TABLE VI
COMPARISON OF ACTION RECOGNITION ACCURACY (%) NTU RGB-D ACTIVITY DATASET

| Method | Data type | Cross subject | Cross view |
|---|---|---|---|
| Skepxel$_{loc+vel}$ [41] | Joints | **81.3** | **89.2** |
| STA-LSTM [44] | Joints | 73.4 | 81.2 |
| ST-LSTM [45] | Joints | 69.2 | 77.7 |
| HPM$_{(RGB+D)}$_Traj [34] | RGB-D | **80.9** | **86.1** |
| HPM_TM+DT [17] | RGB-D | 77.5 | 84.5 |
| Re-TCN [46] | Joints | 74.3 | 83.1 |
| dyadic [47] | RGB-D | 62.1 | - |
| DeepResnet-56 [48] | Joints | 78.2 | 85.6 |
| Motion Stream | RGB | 62 | 68.7 |
| STD stream | Depth Maps | 68.3 | 72.4 |
| Proposed Hybrid Approach | RGB-D (max fusion) | 71.6 | 79.8 |
| | RGB-D (late fusion) | 75.7 | 83 |
| | RGB-D (product fusion) | **79.4** | **84.1** |

TABLE VII
AVERAGE COMPUTATION SPEED (FRAME PER SEC: FPS)

| Method | Training | Testing |
|---|---|---|
| NKTM [32] | 12fps | 16fps |
| HOPC [20] | 0.04fps | 0.5fps |
| HPM+TM [18] | 22fps | 25fps |
| Ours | **36fps** | **28 fps** |

## V. CONCLUSION

We presented a novel two stream RGBD deep framework that capitalizes on view-invariant characteristics of both depth and RGB data streams to make action recognition insensitive to view variations. The proposed approach processes the RGB based motion stream and depth based STD stream independently to exploit the individual modalities without any influence of each other. Motion stream captures the motion details of the action in the form of RGB-Dynamic images (RGB-DIs) which are processed with fine-tuned InceptionV3 deep network. STD stream captures the view-invariant temporal dynamics of depth frames of key poses using HPM [18] model followed by sequence of Bi-LSTM and LSTM layers that helped to learn long-term view-invariant shape dynamics of the actions. Structural Similarity Index Matrix (SSIM) based key pose extraction helps to inspect only major shape variations during the action reducing the redundant frames having minor shape changes. The late fusion of scores of the motion stream and STD stream is used to predict the label of the test sample. To validate the performance of the proposed framework experiments are conducted on three publically available multi-view datasets-NUCLA multi-view dataset, UWA3D II Activity dataset, and NTU RGB-D Activity dataset using cross view and cross-subject cross-validation scheme. The ROC representation of the recognition performance of the proposed framework for each test view exhibits the improved AUC for late fusion over motion and STD streams individually. It is also, noticed that the recognition accuracy of the framework is consistent for different views that confirms the view-invariant characteristics of the framework. In the last, comparisons with other state-of-the-arts are outlined for the proposed deep architecture proving the superiority of the framework in terms of time efficiency and accuracy both.

In future work, we aim to utilize the skeleton details of actions along with depth details for different viewpoints to make the action recognition more robust to intra-class variations of large set of samples and maintaining time efficient performance of the system.

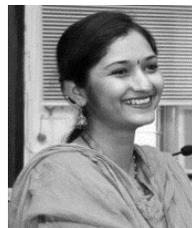
**Chhavi Dhiman** received the B.Tech. from Indira Gandhi Delhi Technical University for Women (IGDTUW), Delhi, India, in 2011 and M.Tech. from Delhi Technological University (DTU), Delhi, India, in 2014. She is currently a Research Scholar and pursuing Ph.D. degree from the Department of Electronics and Communication Engineering, Delhi Technological University, Delhi, India. Her current research interest includes Machine Learning, Deep Learning, Pattern Recognition, Human Action Identification and Classification.

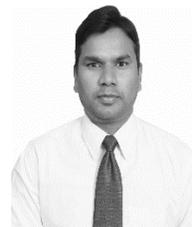
**Dinesh Kumar Vishwakarma (M'16, SM'19)** received the B.Tech. degree from Dr. Ram Manohar Lohia Avadh University, Faizabad, India, in 2002, the M.Tech. degree from the Motilal Nehru National Institute of Technology, Allahabad, India, in 2005, and the Ph.D. degree from Delhi Technological University, New Delhi, India, in 2016. He is currently an Associate Professor with the Department of Information Technology, Delhi Technological University, New Delhi. His current research interests include Computer Vision, Machine Learning, Deep Learning, Sentiment Analysis, Fake News and Rumor Analysis, Crowd Behaviour Analysis, Person Re-Identification, Human Action and Activity Recognition. He is a reviewer of various Journals/Transactions of IEEE, Elsevier, and Springer. He has been awarded with "Premium Research Award" by Delhi Technological University, Delhi, India in 2018.